\documentstyle{article}

\def\nspl{\hbox{\hbox to 20pt{\hrulefill}}}
\def\buttline{\hbox{\hbox to 20pt{\ \hrulefill \ }}}
\newcommand{\mc}{\multicolumn}

\makeatletter

  \newlength\titlebox 
 \twocolumn 

\def\addcontentsline#1#2#3{}

\def\maketitle{\par
 \begingroup
   \def\thefootnote{\fnsymbol{footnote}}
   \def\@makefnmark{\hbox to 0pt{$^{\@thefnmark}$\hss}}
   \twocolumn[\@maketitle] \@thanks
 \endgroup
 \setcounter{footnote}{0}
 \let\maketitle\relax \let\@maketitle\relax
 \gdef\@thanks{}\gdef\@author{}\gdef\@title{}\let\thanks\relax}
\def\@maketitle{\vbox to \titlebox{\hsize\textwidth
 \linewidth\hsize \vskip 0.625in minus 0.125in \centering
 {\Large\bf \@title \par} \vskip 0.2in plus 1fil minus 0.1in
 {\def\and{\unskip\enspace{\rm and}\enspace}%
  \def\And{\end{tabular}\hss \egroup \hskip 1in plus 2fil 
           \hbox to 0pt\bgroup\hss \begin{tabular}[t]{c}\bf}%
  \def\AND{\end{tabular}\hss\egroup \hfil\hfil\egroup
          \vskip 0.25in plus 1fil minus 0.125in
           \hbox to \linewidth\bgroup\large \hfil\hfil
             \hbox to 0pt\bgroup\hss \begin{tabular}[t]{c}\bf}
  \hbox to \linewidth\bgroup\large \hfil\hfil
    \hbox to 0pt\bgroup\hss \begin{tabular}[t]{c}\bf\@author 
                            \end{tabular}\hss\egroup
    \hfil\hfil\egroup}
  \vskip 0.3in plus 2fil minus 0.1in
}}
\renewenvironment{abstract}{\centerline{\large\bf
 Abstract}\vspace{0.5ex}\begin{quote}}{\par\end{quote}\vskip 1ex}

\def\thebibliography#1{\section*{References}
  \global\def\@listi{\leftmargin\leftmargini
               \labelwidth\leftmargini \advance\labelwidth-\labelsep
               \topsep 1pt plus 2pt minus 1pt
               \parsep 0.25ex plus 1pt \itemsep 0.25ex plus 1pt}
  \list {[\arabic{enumi}]}{\settowidth\labelwidth{[#1]}\leftmargin\labelwidth
    \advance\leftmargin\labelsep\usecounter{enumi}}
    \def\newblock{\hskip .11em plus .33em minus -.07em}
    \sloppy
    \sfcode`\.=1000\relax}

\def\@up#1{\raise.2ex\hbox{#1}}

\def\@citex[#1]#2{\if@filesw\immediate\write\@auxout{\string\citation{#2}}\fi
  \def\@citea{}\@cite{\@for\@citeb:=#2\do
     {\@citea\def\@citea{; }\@ifundefined
       {b@\@citeb}{{\bf ?}\@warning
        {Citation `\@citeb' on page \thepage \space undefined}}%
 {\csname b@\@citeb\endcsname}}}{#1}}

\let\@internalcite\cite
\def\cite{\def\citename##1{##1, }\@internalcite}
\def\shortcite{\def\citename##1{}\@internalcite}
\def\newcite{\leavevmode\def\citename##1{{##1} (}\@internalciteb}

\def\@citexb[#1]#2{\if@filesw\immediate\write\@auxout{\string\citation{#2}}\fi
  \def\@citea{}\@newcite{\@for\@citeb:=#2\do
    {\@citea\def\@citea{;\penalty\@m\ }\@ifundefined
       {b@\@citeb}{{\bf ?}\@warning
       {Citation `\@citeb' on page \thepage \space undefined}}%
\hbox{\csname b@\@citeb\endcsname}}}{#1}}
\def\@internalciteb{\@ifnextchar [{\@tempswatrue\@citexb}{\@tempswafalse\@citexb[]}}

\def\@newcite#1#2{{#1\if@tempswa, #2\fi)}}

\def\@biblabel#1{\def\citename##1{##1}[#1]\hfill}

\def\@cite#1#2{({#1\if@tempswa , #2\fi})}

\def\thebibliography#1{\vskip\parskip%
\vskip\baselineskip%
\def\baselinestretch{1}%
\ifx\@currsize\normalsize\@normalsize\else\@currsize\fi%
\vskip-\parskip%
\vskip-\baselineskip%
\section*{References\@mkboth
 {References}{References}}\list
 {}{\setlength{\labelwidth}{0pt}\setlength{\leftmargin}{\parindent}
 \setlength{\itemindent}{-\parindent}}
 \def\newblock{\hskip .11em plus .33em minus -.07em}
 \sloppy\clubpenalty4000\widowpenalty4000
 \sfcode`\.=1000\relax}

\def\thesourcebibliography#1{\vskip\parskip%
\vskip\baselineskip%
\def\baselinestretch{1}%
\ifx\@currsize\normalsize\@normalsize\else\@currsize\fi%
\vskip-\parskip%
\vskip-\baselineskip%
\section*{Sources of Attested Examples\@mkboth
 {Sources of Attested Examples}{Sources of Attested Examples}}\list
 {}{\setlength{\labelwidth}{0pt}\setlength{\leftmargin}{\parindent}
 \setlength{\itemindent}{-\parindent}}
 \def\newblock{\hskip .11em plus .33em minus -.07em}
 \sloppy\clubpenalty4000\widowpenalty4000
 \sfcode`\.=1000\relax}

\def\@lbibitem[#1]#2{\item[]\if@filesw 
      { \def\protect##1{\string ##1\space}\immediate
        \write\@auxout{\string\bibcite{#2}{#1}}\fi\ignorespaces}}

\def\@bibitem#1{\item\if@filesw \immediate\write\@auxout
       {\string\bibcite{#1}{\the\c@enumi}}\fi\ignorespaces}

\def\section{\@startsection {section}{1}{\z@}{-2.0ex plus
    -0.5ex minus -.2ex}{1.5ex plus 0.3ex minus .2ex}{\large\bf\raggedright}}
\def\subsection{\@startsection{subsection}{2}{\z@}{-1.8ex plus
    -0.5ex minus -.2ex}{0.8ex plus .2ex}{\normalsize\bf\raggedright}}
\def\subsubsection{\@startsection{subsubsection}{3}{\z@}{1.5ex plus
   0.5ex minus .2ex}{0.5ex plus .2ex}{\normalsize\bf\raggedright}}
\def\paragraph{\@startsection{paragraph}{4}{\z@}{1.5ex plus
   0.5ex minus .2ex}{-1em}{\normalsize\bf}}
\def\subparagraph{\@startsection{subparagraph}{5}{\parindent}{1.5ex plus
   0.5ex minus .2ex}{-1em}{\normalsize\bf}}

\skip\footins 9pt plus 4pt minus 2pt
\def\footnoterule{\kern-3pt \hrule width 5pc \kern 2.6pt }
\labelwidth\leftmargini\advance\labelwidth-\labelsep \labelsep 5pt

\def\@listi{\leftmargin\leftmargini}
\def\@listii{\leftmargin\leftmarginii
   \labelwidth\leftmarginii\advance\labelwidth-\labelsep
   \topsep 2pt plus 1pt minus 0.5pt
   \parsep 1pt plus 0.5pt minus 0.5pt
   \itemsep \parsep}
\def\@listiii{\leftmargin\leftmarginiii
    \labelwidth\leftmarginiii\advance\labelwidth-\labelsep
    \topsep 1pt plus 0.5pt minus 0.5pt 
    \parsep \z@ \partopsep 0.5pt plus 0pt minus 0.5pt
    \itemsep \topsep}
\def\@listiv{\leftmargin\leftmarginiv
     \labelwidth\leftmarginiv\advance\labelwidth-\labelsep}
\def\@listv{\leftmargin\leftmarginv
     \labelwidth\leftmarginv\advance\labelwidth-\labelsep}
\def\@listvi{\leftmargin\leftmarginvi
     \labelwidth\leftmarginvi\advance\labelwidth-\labelsep}

\belowdisplayskip \abovedisplayskip
\def\@normalsize{\@setsize\normalsize{11pt}\xpt\@xpt}
\def\small{\@setsize\small{10pt}\ixpt\@ixpt}
\def\footnotesize{\@setsize\footnotesize{10pt}\ixpt\@ixpt}
\def\scriptsize{\@setsize\scriptsize{8pt}\viipt\@viipt}
\def\tiny{\@setsize\tiny{7pt}\vipt\@vipt}
\def\large{\@setsize\large{14pt}\xiipt\@xiipt}
\def\Large{\@setsize\Large{16pt}\xivpt\@xivpt}
\def\LARGE{\@setsize\LARGE{20pt}\xviipt\@xviipt}
\def\huge{\@setsize\huge{23pt}\xxpt\@xxpt}
\def\Huge{\@setsize\Huge{28pt}\xxvpt\@xxvpt}

\makeatother

\makeatletter

\def\pscmd#1{\special{ps:@beginspec
/nodemargin \@int{\the\nodemargin}\space pt def %
\@int{\the\treelinewidth}\space pt setlinewidth %
\ifdim\dashlength=0pt [] 0 setdash%
/arrowwidth \@int{\the\arrowwidth}\space pt def %
/arrowlength \@int{\the\arrowlength}\space pt def %
\else [\@int{\the\dashlength}\space pt] 0 setdash \fi\space
#1 %
@endspec}}

\newdimen\nodemargin %
\newdimen\treelinewidth
\newdimen\dashlength

\newdimen\arrowwidth
\newdimen\arrowlength
{\catcode`p=12 \catcode`t=12 \gdef\wnum#1pt{#1}}
\def\@int#1{\expandafter\wnum#1}

\def\node#1#2{\leavevmode
              \setbox1=\hbox{#2}\pscmd{/#1 \@int{\the\wd1} \space pt
                                           \@int{\the\ht1} \space pt
                                           \@int{\the\dp1} \space pt
                                       node}\box1\relax}

\def\nodepoint#1{\@ifnextchar [{\@nodepoint{#1}}{\@nodepoint{#1}[0pt][0pt]}}

\def\@nodepoint#1[#2][#3]{{\@tempdima=#2 \@tempdimb=#3%
\pscmd{/nodemargin 0 def %
/#1 \@int{\the\@tempdima}\space pt \@int{\the\@tempdimb}\space pt %
rmoveto 0 0 0 node}}}

\newtoks\pos@t   \pos@t={top}    
\newtoks\pos@b   \pos@b={bottom} 
\newtoks\pos@l   \pos@l={left}   
\newtoks\pos@r   \pos@r={right}  
\newtoks\pos@tl   \pos@tl={topleft}    
\newtoks\pos@tr   \pos@tr={topright}   
\newtoks\pos@bl   \pos@bl={bottomleft} 
\newtoks\pos@br   \pos@br={bottomright}

\def\nodeconnect{\@ifnextchar [{\@nodeconnect}{\@nodeconnect[b]}}
\def\@nodeconnect[#1]#2{\@ifnextchar [{\@@nodeconnect[#1]{#2}}{\@@nodeconnect[#1]{#2}[t]}}

\def\@@nodeconnect[#1]#2[#3]#4{\pscmd{%
/#2 getnode node\the\csname pos@#1\endcsname \space%
/#4 getnode node\the\csname pos@#3\endcsname\space nodeconnect}}

\def\nodecurve{\@ifnextchar [{\@nodecurve}{\@nodecurve[b]}}
\def\@nodecurve[#1]#2{\@ifnextchar [{\@@nodecurve[#1]{#2}}{\@@nodecurve[#1]{#2}[t]}}

\def\@@nodecurve[#1]#2[#3]#4#5{\@ifnextchar [{\@@@nodecurve[#1]{#2}[#3]{#4}{#5}}%
{\@@@nodecurve[#1]{#2}[#3]{#4}{#5}[#5]}}

\def\@@@nodecurve[#1]#2[#3]#4#5[#6]{\@tempdima=#5%
\@tempdimb=#6%
\pscmd{/depth \@int{\the\@tempdimb} \space pt def %
 /#4 \the\csname pos@#3\endcsname cur\space %
 /depth \@int{\the\@tempdima} \space pt def %
 /#2 \the\csname pos@#1\endcsname cur\space %
nodecurve}}

\newif\iftransparent

\def\nodetriangle#1#2{\pscmd{/#1 /#2 nodetriangle}}

\makeatother

\makeatletter

\newcounter{enums}

\newdimen\widelabel
\def\enumsentence{\@ifnextchar[{\@enumsentence}
{\refstepcounter{enums}\@enumsentence[(\theenums)]}}

\long\def\@enumsentence[#1]#2{\begin{list}{}{%
\advance\leftmargin by\widelabel \advance\labelwidth by \widelabel}
\item[#1] #2
\end{list}}

\newcounter{tempcnt}

\newcommand{\ex}[1]{\setcounter{tempcnt}{\value{enums}}%
\addtocounter{tempcnt}{#1}%
\arabic{tempcnt}}

\def\@item[#1]{\if@noparitem \@donoparitem
  \else \if@inlabel \indent \par \fi
         \ifhmode \unskip\unskip \par \fi 
         \if@newlist \if@nobreak \@nbitem \else
                        \addpenalty\@beginparpenalty
                        \addvspace\@topsep \addvspace{-\parskip}\fi
           \else \addpenalty\@itempenalty \addvspace\itemsep 
          \fi 
    \global\@inlabeltrue 
\fi
\everypar{\global\@minipagefalse\global\@newlistfalse 
          \if@inlabel\global\@inlabelfalse \hskip -\parindent \box\@labels
             \penalty\z@ \fi
          \everypar{}}\global\@nobreakfalse
\if@noitemarg \@noitemargfalse \if@nmbrlist \refstepcounter{\@listctr}\fi \fi
\setbox\@tempboxa\hbox{\makelabel{#1}}%
\global\setbox\@labels
 \hbox{\unhbox\@labels \hskip \itemindent
       \hskip -\labelwidth \hskip -\labelsep 
       \ifdim \wd\@tempboxa >\labelwidth 
                \box\@tempboxa
          \else \hbox to\labelwidth {\unhbox\@tempboxa}\fi
       \hskip \labelsep}\ignorespaces}

\newcounter{enumsi}

\newdimen\eeindent
\def\@mklab#1{\hfil#1}
\def\enummklab#1{\hfil(\eelabel)\hbox to \eeindent{\hfil#1}}
\def\enummakelabel#1{\enummklab{#1}\global\let\makelabel=\@mklab}
\def\toplabel#1{{\edef\@currentlabel{\p@enums\theenums}\label{#1}}}

\def\eenumsentence{\@ifnextchar[{\@eenumsentence}
{\refstepcounter{enums}\@eenumsentence[\theenums]}}

\long\def\@eenumsentence[#1]#2{\def\eelabel{#1}\let\holdlabel\makelabel%
\begin{list}{\alph{enumsi}.}{\usecounter{enumsi}%
\advance\leftmargin by \eeindent \advance\leftmargin by \widelabel%
\advance\labelwidth by \eeindent \advance\labelwidth by \widelabel%
\let\makelabel=\enummakelabel}
#2
\end{list}\let\makelabel\holdlabel}

\def\shortex#1#2#3#4{\begin{tabular}[t]{@{}*{#1}{l@{\ }}}
#2\\ #3\\ \multicolumn{#1}{@{}l@{}}{\parbox{\linewidth}{#4}}
\end{tabular}}

\makeatother

\newcommand{\bt}{\begin{tabular}}
\newcommand{\et}{\end{tabular}}
\newcommand{\st}{\shortex}
\newcommand{\ncon}{\nodeconnect}
\newcommand{\es}{\enumsentence}
\newcommand{\ees}{\eenumsentence}
\newcommand{\ua}{\uparrow}
\newcommand{\da}{\downarrow}
\newcommand{\ra}{\rightarrow}

\title{\vspace{-0.5in}Syntactic Analyses for Parallel Grammars:
  Auxiliaries and Genitive NPs}

\author{Miriam Butt \and Christian Fortmann \and Christian Rohrer \\
Institut f\"{u}r Maschinelle Sprachverarbeitung \\ Universit\"{a}t
Stuttgart \\ 
Azenbergstr. 12 \\ 70174 Stuttgart, Germany \\
  \{mutt$|$fortmann$|$rohrer\}@ims.uni-stuttgart.de \\}

\begin{document}

\maketitle
\vspace{-0.5in}
\begin{abstract}

  This paper focuses on two disparate aspects of German syntax from
  the perspective of parallel grammar development.  As part of a
  cooperative project, we present an innovative approach to
  auxiliaries and multiple genitive NPs in German.  The LFG-based
  implementation presented here avoids unnessary structural complexity
  in the representation of auxiliaries by challenging the traditional
  analysis of auxiliaries as raising verbs.  The approach developed
  for multiple genitive NPs provides a more abstract, language
  independent representation of genitives associated with nominalized
  verbs.  Taken together, the two approaches represent a step towards
  providing uniformly applicable treatments for differing languages,
  thus lightening the burden for machine translation.
\end{abstract}

\section{Introduction}

Within the cooperative parallel grammar project {\sc pargram}
(IMS-Stuttgart, Xerox-Palo Alto, Xerox-Grenoble), the analysis and
representation of structures in the grammars must be viewed from a
more global perspective than that of the individual languages (German,
English, French).  One major goal of {\sc pargram} is the development
of broad coverage grammars which are also modular and easy to
maintain.  Another major goal is the construction of {\em parallel}
analyses for sentences of the same type in German, English, and
French.  If this can be achieved, the problem faced by machine
translation (MT) could be greatly reduced.  Due to the recent
development of a faster and more powerful version of the LFG
(Lexical-Functional-Grammar) based {\em Grammar Writer's Workbench}
(Kaplan and Maxwell 1993) at Xerox, the implementation of a
linguistically adequate, broad coverage grammar appears viable.  Given
the flexible projection-based architecture of LFG (Dalrymple et
al.~1995) and the MT approach presented in Kaplan et
al.~(1989),\footnote{See also Sadler et al.~(1990), Sadler and
  Thompson (1991), Kaplan and Wedekind (1993), Butt (1994) for further
  work on MT within LFG.} a robust MT system is already in place.

In this paper, we concentrate on two issues within the broader
perspective of {\sc pargram}: the treatment of auxiliaries and the
transparent representation of multiple genitive NPs in German.  These
phenomena represent two areas for which generally accepted proposals
exist, but whose implementation in the context of parallel grammar
development throws up questions as to their wider, crosslinguistic,
feasibility.  With respect to auxiliaries, the standard {\em raising}
approach that is usually adopted yields undesirable structural
complexity and results in idiosyncratic, language particular analyses
of the role of auxiliaries.  With regard to genitive NPs, the standard
analysis for German yields structures which are too ambiguous for a
succesful application of machine translation.  The following sections
present a solution in that morphological wellformedness
conditions are stated at a separate component, the {\em morphology
  projection}. Furthermore, a representation of argument structure is
implemented that is related to, but not identical to the
representation of grammatical functions.  Language particular
idiosyncratic requirements are thus separated out from the language
universal information required for further semantic interpretation, or
machine translation.

\section{The Formalism}

The architecture of LFG assumed here is the ``traditional''
architecture described in Bresnan (1982), as well as the newer
advances within LFG (Dalrymple et al.~, 1995). A grammar is viewed as
a set of {\em correspondences} expressed in terms of {\em projections}
from one level of representation to another.  Two fundamental levels
of representations within LFG are the c(onstitutent)-structure and the
f(unctional)-structure.  The c-structure encodes idiosyncratic phrase
structural properties of a given language, while the f-structure
provides a language universal representation of grammatical functions
(e.g., {\sc subj}ect, {\sc obj}ect), complementation, tense, binding,
etc.  The correspondence between c-structure and f-structure is
not onto or one-to-one, but many-to-one, allowing an abstraction over
idiosyncratic c-structure properties of a language (e.g.,
discontinuous constituents).

In addition, several proposals exploring possible representations of a
s(emantic)-structure have been made over the years (e.g. Halvorsen and
Kaplan (1988), Dalrymple et al.~ (1993)).  As the 
realization of a separate semantic component is only planned for the
latter stages within {\sc pargram}, no further discussion of possible
formalisms will take place here.  It should be noted, however, that
rudimentary semantic information, such as argument structure
information (lexical semantics), is encoded within the f-structures in
order to facilitate transfer in some cases.  A case in point is
presented in the section on German genitive NPs.  

\section{Auxiliaries --- a flat approach}

\subsection{The Received Wisdom}

Auxiliaries have given rise to lively debates concerning their exact
syntactic status (e.g.~Chomsky (1957), Ross (1967), Pullum and Wilson
(1977), Akmajian et al.~(1979), Gazdar et al.~(1982)): are they simply
main verbs with special properties, or should they instantiate a
special category {\sc aux}?  Within current lexical approaches
(Lexical-Functional-Grammar (LFG), Head-driven Phrase Structure
Grammar (HPSG)), auxiliaries (e.g.~{\em have}, {\em be}) and modals
(e.g.~{\em must}, {\em should}) are treated as {\em raising} verbs,
which are marked as special in some way: in HPSG through an [{\sc
aux}: +] feature (Pollard and Sag 1994), in LFG (Bresnan 1982) by a
difference in {\sc pred} value.\footnote{See Falk (1984) for an early
LFG treatment of `do' in line with that proposed here, and Abeill\'{e}
and Godard (1994) for a similar treatment in French.}  However, newer
work within LFG (Bresnan 1995, T.H.~King 1995) has been moving away
from the raising approach towards an analysis where auxiliaries are
elements which contribute to the clause only tense/aspect, agreement,
or voice information, but not a subcategorization frame.  This view is
also in line with approaches within GB (Government-Binding), which see
auxiliaries simply as possible instantiations of the functional
category I (see also Halle and Marantz (1993)). 

The ``traditional'' treatment of auxiliaries in both HPSG (Pollard and Sag
1994) and LFG has its roots in Ross's (1967) proposal to treat
auxiliaries and modals on a par with main verbs.\footnote{The term
{\em auxiliary} has often been taken to subsume both modals and
elements such as {\em have} and {\em be}.  However, the distinction
between the two is necessary not only semantically, but also
syntactically.  In German and (some dialects of)
English modals can be stacked, while the distribution of auxiliaries
is more restricted.  Also, assuming that semantic interpretation is
driven primarily off of the f-structure, the relative embedding of
modals must be preserved at that level in order to allow an
interpretation of their scope and semantic force.}\ In particular,
auxiliaries are treated as a subclass of raising verbs (e.g.~Pollard
and Sag (1994), Falk (1984)).  For example, a simple sentence like (\ex{1})
would correspond to the c-structure and f-structure shown in (\ex{2})
and (\ex{3}), respectively. Note that the level of embedding in the
f-structure exactly mirrors the c-structure: each verbal element takes
a complement.  

\es{\st{7}
{Der & Fahrer & wird & den & Hebel & gedreht & haben}
{the & driver & will & the & lever & turned & gave}
{`The driver will have turned the lever.'}}

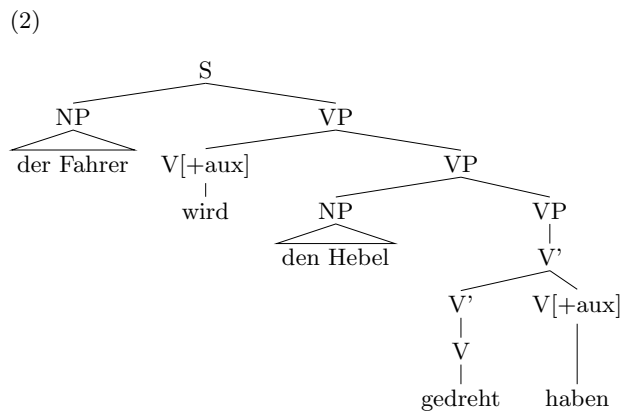
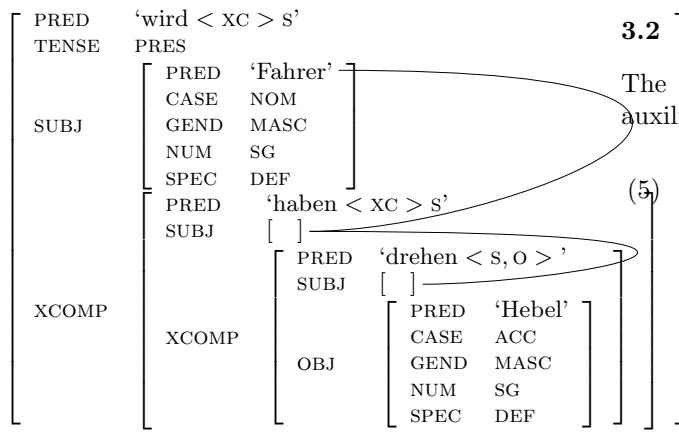
\small
\vbox{
\es{\ }
\bt[t]{ccccccc}
& \node{z}{S} & \\[2ex]
\node{b}{NP} & &  \node{a}{VP} \\ [2ex]
\node{c}{der Fahrer} &   \node{d}{V[+aux]} & & \node{e}{VP}  \\ [2ex]
& \node{g}{wird} & \node{f}{NP} & & \mc{3}{c}{\node{v}{VP}}  \\ [2ex]
& & \node{h}{den Hebel} & & \mc{3}{c}{\node{i}{V'}} \\ [2ex]
& & & \node{j}{V'} &  \node{k}{V[+aux]} \\ [2ex]
& & & \node{u}{V} \\ [2ex]
& & &  \node{l}{gedreht} &  \node{m}{haben} \\
\et
\ncon{z}{a} \ncon{z}{b}
\nodetriangle{b}{c} \ncon{a}{d} \ncon{a}{e}
\ncon{d}{g} \ncon{e}{f} \ncon{e}{v}
\nodetriangle{f}{h} \ncon{v}{i} 
\ncon{i}{j} \ncon{i}{k} 
\ncon{j}{u} \ncon{k}{m} 
\ncon{u}{l}
}
\normalsize

\footnotesize
\vbox{
\es{\ }
\[ \left[  \begin{array}{ll}
\hbox{{\sc pred}} & \hbox{`wird}<\hbox{{\sc xc}}>\hbox{{\sc s}}\mbox{'} \\ 
\hbox{{\sc tense}} & \hbox{{\sc pres}} \\
\hbox{{\sc subj}} & \left[  \begin{array}{ll} 
\hbox{{\sc pred}} & \node{a}{\hbox{`Fahrer'}} \\
\hbox{{\sc case}} & \hbox{{\sc nom}} \\
\hbox{{\sc gend}} & \hbox{{\sc masc}} \\
\hbox{{\sc num}} & \hbox{{\sc sg}} \\
\hbox{{\sc spec}} & \hbox{{\sc def}} \\
                 \end{array} \right]  \\

\hbox{{\sc xcomp}} & \left[ \begin{array}{ll} \hbox{{\sc pred}} &
`\hbox{haben}<\hbox{{\sc xc}}>\hbox{{\sc s}}\mbox{'} \\   

\hbox{{\sc subj}} & \node{b}{[\ \ \ ]} \\ 

\hbox{{\sc xcomp}} & \left[ \begin{array}{ll} \hbox{{\sc pred}} &
`\hbox{drehen}<\hbox{{\sc s}}, \hbox{{\sc o}}>\mbox{'} \\   

\hbox{{\sc subj}} & \node{c}{[\ \ \ ]} \\ 

\hbox{{\sc obj}} & \left[ \begin{array}{ll} 
\hbox{{\sc pred}} & \hbox{`Hebel'} \\
\hbox{{\sc case}} & \hbox{{\sc acc}} \\
\hbox{{\sc gend}} & \hbox{{\sc masc}} \\
\hbox{{\sc num}} & \hbox{{\sc sg}} \\
\hbox{{\sc spec}} & \hbox{{\sc def}} \\
                 \end{array} \right] 

\end{array} \right]
\end{array} \right]
\end{array} \right] \]
\nodecurve[r]{a}[r]{b}{3in}[1in]
\nodecurve[r]{b}[r]{c}{3in}[.7in]
}

\normalsize

The main reasons to treat auxiliaries as complement taking verbs in
English are: 1) an account of VP-ellipsis, VP-topicalization, etc.\ 
follows immediately; 2) restrictions on the nature of the verbal
complement (progressive, past participle, etc.) following the
auxiliary can be stated straightforwardly (Pullum and Wilson (1977),
Akmajian et al.~(1979), Gazdar et al.~(1982)).  The latter point holds
for German as well, and in fact, without some sort of a hierarchical
structure, stating wellformedness conditions on a string of multiple
auxiliaries becomes wellnigh impossible in light of the greater
ordering possibilities granted by the flexible German word order.
There are also major reasons, however, for not adopting this analysis:
1) linguistic adequacy; 2) unmotivated structural complexity; 3)
non-parallel analyses for predicationally equivalent sentences.
Consider the French equivalent of (\ex{-2}) in (\ex{1}).

\es{\st{6}
{Le & conducteur & aura & tourn\'{e} & le & levier}
{the & driver & will have & turned & the & lever}
{`The driver will have turned the lever.'}}

As argued by Akmajian et al.~(1979), crosslinguistic evidence
indicates that elements bearing only tense, mood, or voice should
belong to a distinct syntactic category.  In many languages, like
French or Japanese, the information carried by {\em will} (future), or
{\em have} (perfect) is realized morphologically rather than
periphrastically.  The analysis in (\ex{0}) thus effectively claims
that there exists a deep difference in the predicational structure of
auxiliaries like {\em will} and {\em have} and the French {\em
  aura}.\footnote{Note that {\em wird} `will' is often analyzed as a
  modal in accordance with Vater (1975).  However, the arguments
  presented there are not conclusive.} This is not desirable from a
crosslinguistic point of view, nor is it helpful for MT.  

\subsection{Alternative Implementation}

The approach adopted here is a {\em flat} analysis of auxiliaries at
f-structure ((\ex{1})).

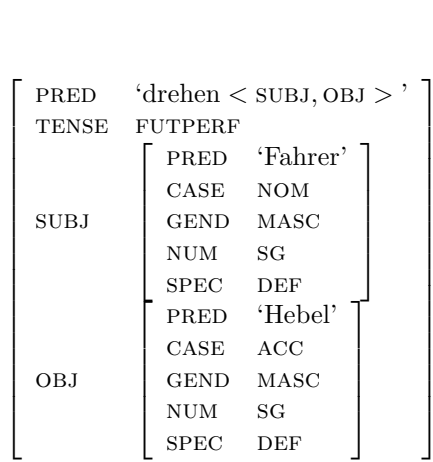
\es{
\[ \left[  \begin{array}{ll}

\hbox{{\sc pred}} & `\hbox{drehen}<\hbox{{\sc subj}}, \hbox{{\sc obj}} >\mbox{'} \\   
\hbox{{\sc tense}} & \hbox{{\sc futperf}} \\
\hbox{{\sc subj}} & \left[  \begin{array}{ll} 
\hbox{{\sc pred}} & \node{a}{\hbox{`Fahrer'}} \\
\hbox{{\sc case}} & \hbox{{\sc nom}} \\
\hbox{{\sc gend}} & \hbox{{\sc masc}} \\
\hbox{{\sc num}} & \hbox{{\sc sg}} \\
\hbox{{\sc spec}} & \hbox{{\sc def}} \\
                 \end{array} \right] \\

\hbox{{\sc obj}} & \left[ \begin{array}{ll} 
\hbox{{\sc pred}} & \hbox{`Hebel'} \\
\hbox{{\sc case}} & \hbox{{\sc acc}} \\
\hbox{{\sc gend}} & \hbox{{\sc masc}} \\
\hbox{{\sc num}} & \hbox{{\sc sg}} \\
\hbox{{\sc spec}} & \hbox{{\sc def}} \\
                 \end{array} \right] \\

\end{array} \right] \]
}

The auxiliaries {\em wird} `will' and {\em haben} `have' now only
contribute information as to the overall tense, but do not
subcategorize for complements.  Structural phenomena like VP-ellipsis,
coordination, or topicalization can, however, still be accounted for
in terms of an appropriate embedding at c-structure (cf.~(\ex{-3})).
The role of auxiliaries in natural language is now adequately modeled,
in particular with respect to a more realistic treatment of tense
(compare (\ex{-2}) and (\ex{0})), as the French (\ex{-1}) has
essentially the same f-structure as (\ex{0}).\footnote{The
  construction of the value for the composed tenses results from a
  complex interaction between the lexical entries. Note that this
  treatment does not as yet include a fine-grained represention of
  tense and aspect.  This is the subject of ongoing work.  The
  treatment presented here provides the basis needed for a thorough
  crosslinguistic analysis of temporal and aspectual phenomena.}

However, the flat f-structure in (\ex{0}) provides no room for a
statement of selectional requirements, allowing massive overgeneration
(e.g.~nothing blocks the presence of two {\em haben} in (\ex{-4})).
Neither can the particular order of auxiliaries be regulated.  Our
solution takes advantage of LFG's flexible projection-based
architecture by implementing a projection which models the
hierarchical selectional requirements of auxiliaries, yet does not
interfere with the subcategorizational properties of verbs, as would
be the case under a raising analysis.

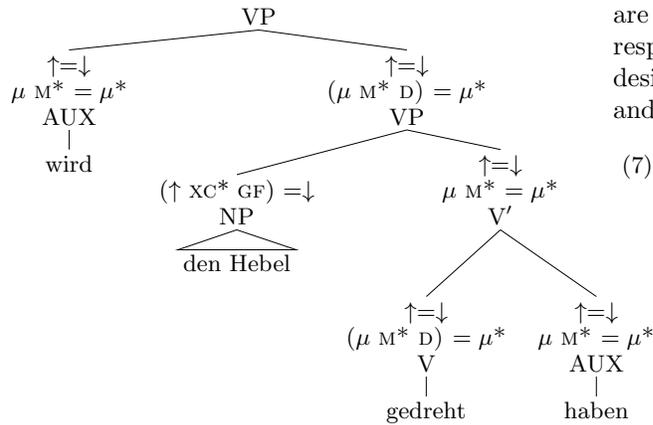
\footnotesize
\vbox{
\es{\ }
\bt[t]{ccccc}
\multicolumn{3}{c}{\node{e}{VP}} \\ [2ex]
\node{f}{$\ua = \da$} &\multicolumn{3}{c}{\node{g}{$\ua = \da$}} \\
 \node{h}{$\mu$ \sc{m}* $=\mu$*} &\multicolumn{3}{c}{\node{i}{($\mu$ \sc{m}* {\sc d}) $=\mu$*}} \\
 \node{j}{AUX} &\multicolumn{3}{c}{\node{k}{VP}} \\ [2ex]
 \node{jj}{wird} & &\multicolumn{2}{c}{\node{m}{$\ua = \da$}} \\
 & \node{l}{($\ua$ {\sc xc}* {\sc gf}) $= \da$} &\multicolumn{2}{c}{\node{n}{$\mu$ \sc{m}* $=\mu$*}} \\
 & \node{v}{NP} &\multicolumn{2}{c}{\node{o}{V$'$}} \\ [2ex]
 & \node{vv}{den Hebel} \\ [2ex]
 & &  \node{p}{$\ua = \da$} & \node{q}{$\ua = \da$} \\
 & & \node{r}{($\mu$ \sc{m}* {\sc d}) $=\mu$*} & \node{s}{$\mu$ \sc{m}* $=\mu$*}  \\
 & & \node{t}{V} & \node{u}{AUX} \\ [2ex]
 & & \node{tt}{gedreht} & \node{uu}{haben} \\
\et

\ncon{e}{f} \ncon{e}{g}
\ncon{j}{jj} \ncon{k}{l} \ncon{k}{m}
\nodetriangle{v}{vv} \ncon{o}{p} \ncon{o}{q}
\ncon{t}{tt} \ncon{u}{uu}
}
\normalsize

In LFG, the flexible word order of German is handled via {\em
  functional uncertainty}, which characterizes long-distance
dependencies without resorting to movement analyses (Netter (1988),
Zaenen and Kaplan (1995)).  As in (\ex{0}), which illustrates our
alternative solution, functional uncertainty is represented by the
Kleene Star (*).\footnote{For space reasons, the {\sc xc} indicates
  {\sc xcomp}, the {\sc d} a {\sc dep}.}\ The annotation on the NPs
indicates that they could fulfill the role of any possible grammatical
function (GF), e.g.~{\sc subj} or {\sc obj}, and that the level of
embedding ranges from zero to infinite.  With every auxiliary
subcategorizing for an {\sc xcomp}, the two NPs could conceivably be
arguments of three different verbs: {\em wird}, {\em haben}, or {\em
  gedreht}.  Thus, the greater structural complexity unnecessarily
increases the search space for the determination of a verb's
arguments. In (\ex{0}), however, the m-structure is projected from the
c-structure parallel to the f-structure through annotations similar to
the usual f-structure annotations.\footnote{The annotation $\mu$ M* in
  (\ex{0}) refers to the m-structure associated with the parent
  c-structure node, and $\mu$* refers to the m-structure associated
  with the daughter node.  The more familiar $\uparrow$ and
  $\downarrow$ of LFG are simply shorthand notations of the same idea,
  but restricted to the projection from c-structure to f-structure:
  $\uparrow = \phi$ \sc{m}*, $\downarrow = \phi$ *.}\ Statements about
``morphological'' dependents ({\sc dep}) are thus decoupled from
functional uncertainty: the relation of NP arguments to their
predicator now does not extend through various layers of artificial
structural complexity ({\sc xcomp}s).  For VP-topicalization or
extraposition an unbounded long-distance dependency must still be
assumed.  However, as the functional uncertainty path for auxiliaries
is distributed only over the m-structure of the verb complex (($\mu
\ua$ {\sc dep}*) $=\mu \da$), and does not involve the resolution of
the role of NP arguments, there are in fact differing paths of
functional uncertainty involved.  The dependencies between predicators
and their arguments and auxiliaries and their dependents are thus
neatly factored out. The m-structure corresponding to the matrix VP in
(\ex{0}) is (\ex{1}).  The desired flat f-structure resulting from the
usual $\ua$ and $\da$ annotations is as in (\ex{-1}).

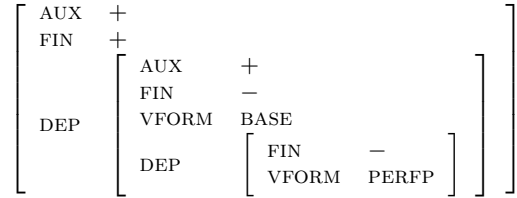
\small
\es{\ 
\[ \left[  \begin{array}{ll}

\hbox{{\sc aux}} & + \\
\hbox{{\sc fin}} & + \\

\hbox{{\sc dep}} & \left[ \begin{array}{ll} 

\hbox{{\sc aux}} & + \\
\hbox{{\sc fin}} & - \\
\hbox{{\sc vform}} & \hbox{{\sc base}} \\

\hbox{{\sc dep}} & \left[ \begin{array}{ll} 
\hbox{{\sc fin}} & - \\
\hbox{{\sc vform}} & \hbox{{\sc perfp}} \\
                 \end{array} \right]  \\ 
                 \end{array} \right] \\

\end{array} \right] \]
}

\normalsize

Like the f-structure, the m-structure is an attribute-value matrix.
It encodes language-specific information about idiosyncratic
constraints on morphological forms.  The m-structure is not derived
from the f-structure.  Rather, both representations are in
simultaneous correspondence with the c-structure.  The following
(abbreviated) lexical entry exemplifies the pieces of information
needed.  The disjunctive lexical entry for {\em wird} `will' in
(\ex{1}) takes the various combinatory possibilities of auxiliaries
and main verbs into account, and provides the appropriate tense
feature.  For example, it requires that the embedded {\sc vform} be {\sc base},
and that there be no passive involved for a simple future like {\em
  wird drehen}.  

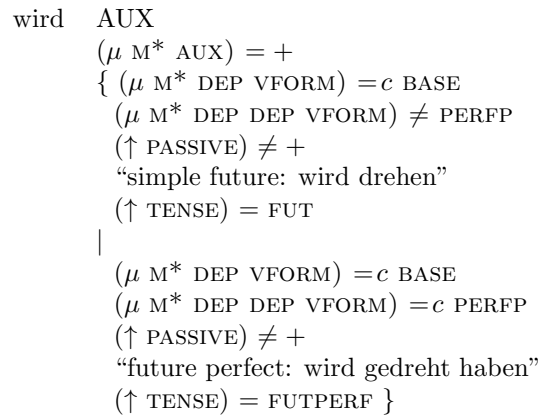
\es{\bt[t]{ll}
wird & AUX  \\
& ($\mu$ \sc{m}* {\sc aux}) $= +$ \\
& \{ ($\mu$ \sc{m}* {\sc dep vform}) $=\!c$ {\sc base} \\
& \ \  ($\mu$ \sc{m}* {\sc dep dep vform}) $\neq$ {\sc perfp} \\  
& \ \ ($\ua$ {\sc passive}) $\neq + $ \\
& \ \  ``simple future: wird drehen'' \\
& \ \  ($\ua$ {\sc tense}) = {\sc fut} \\
&   $|$ \\
& \ \ ($\mu$ \sc{m}* {\sc dep vform}) $=\!c$ {\sc base} \\
& \ \ ($\mu$ \sc{m}* {\sc dep dep vform}) $=\!c$ {\sc perfp} \\  
& \  \ ($\ua$ {\sc passive}) $\neq + $  \\
& \  \  ``future perfect: wird gedreht haben'' \\
& \ \  ($\ua$ {\sc tense}) $=$ {\sc futperf} \} \\
\et}

Features needed only to ensure language particular wellformedness are
no longer unified into the f-structure, cluttering a representation
that is meant to be language independent.  In our analysis, only
features needed for further semantic interpretation, MT, or for the
expression of language universal syntactic generalizations are
represented at f-structure.  For example, morphologically encoded
information like case, gender, or agreement is needed for statements
as to binding, predicate-argument relations, or the determination of
complex clause structures (given that agreement is generally
clause-bounded), and is therefore represented at f-structure.
Wellformedness conditions on adjective inflection or relative pronoun
agreement, however, can now be stated on the m-structure as
idiosyncratic, language particular information which can be ignored
for purposes of MT or semantic interpretation.

\section{Multiple Genitive NPs}

The differing surface realization of genitives within NPs in English
(preverbal NPs, postverbal PPs), French (postverbal PPs), and German
(preverbal NPs, postverbal PPs or NPs), poses a particular challenge
for a parallel grammar development project like {\sc pargram}.  In
this paper, we suggest a treatment of multiple genitive NPs which not
only accounts for some restrictions on their distribution within
German, but also allows a language independent (universal)
representation, thus facilitating MT.

In general, the distribution of multiple NPs within NPs is an area of
German syntax which has not received a satisfactory account to date
(e.g., Pollard and Sag (1994), Bhatt (1990), Haider (1988)). In
German, nouns generally have at most one genitive which may occur in
a prenominal or postnominal position adjacent to the noun. Both kinds of
genitives have the same morphological shape. However, nominalizations
that are derived from a transitive verb allow for two genitives, one
in the prenominal, the other in the postnominal position.

The function of a genitive is generally expressed as indicating a
possessor: {\sc poss} within LFG. However, in the case of two
genitives, the assignment of two {\sc poss} values violates the
uniqueness-condition on f-structures and is furthermore insufficient
to distinguish the two differing kinds of genitives.  We therefore
propose the utilization of two functions named {\sc gen1} and {\sc
  gen2} in order to avoid association with any specific semantic role.
Furthermore, as genitives in the NP are generally optional, they are
taken to express no governed functions, i.e., they are not
subcategorized for by the noun.  So {\sc gen1} and {\sc gen2} are 
semantic functions in LFG on a par with, say, adjuncts. The
NP rule for German then is (\ex{1}).\footnote{Abstracting away from
  bar-level considerations and further optional constituents, this
  rule captures the restrictions that determine the dislocation of a
  genitive in the matrix NP.}

\es{\bt[t]{lll}
NP $\ra$ & (\{DET:&  $\ua = \da$ \\
&                 $|$ NP: &  ($\ua$ {\sc gen1}) $ = \da$ \}) \\
&                 N: & $\ua = \da$ \\
                & (NP: & ($\ua$ {\sc gen2}) $ = \da$)  \\
\et
}

If the head-noun is not derived from, say, a verb, the single genitive
in either position is interpreted as a possessor. In case of a derived
nominal, however, a genitive is interpreted according to the thematic roles
assigned to the arguments of the verbal base. That means the functions
{\sc gen1} and {sc\ gen2} have to be linked to the appropriate roles.
Neither of the two functions is in principle restricted to any
specific role. But if both genitives are present they must be 
interpreted according to a thematic role hierarchy.

As (\ex{1}) shows, if only one genitive is present, its prenominal
interpretation may be as agent or as patient. A postnominal (single)
genitive is interpreted as agent if the head noun is derived from an
intransitive, and as a patient/theme if derived from a transitive.

\ees{\item[a.]\st{2}
{Elisabeths & Lachen}
{Elisabeth-Gen & laughing}
{`Elisabeth's laughter'}

\item[b.]\st{2}
{Roms & Belagerung}
{Rome-Gen & siege}
{`Rome's siege'}}

However, if two genitives occur, as in (\ex{1}), the prenominal
genitive is restricted to an agent, and the postnominal one to patient. 
This restriction must be encoded at some level, but does not follow
from the distiction between {\sc gen1} and {\sc gen2}, which are functions
that do not bear any semantic content on their own.

\es{\st{3}
{Karls & Behandlung & Peters}
{Karl-Gen & treatment & Peter-Gen}
{`Karl's treatment of Peter'}}

To obtain the correct linking, the argument structure of the verbal
base must be available.  Since MT is based on f-structures within {\sc
  pargram}, the argument structure has to be present at this level of
representation.\footnote{If a semantic or argument projection is
  assumed (e.g., Halvorsen and Kaplan, 1988),  this information may be
  represented at another independent projection.}\ Nominalization is therefore
implemented as a morphologically driven process (lexical rule) which
eliminates {\sc subj} and {\sc obj} from the verb's subcategorization
frame and enters the verb's argument structure into the lexical entry
of the noun.  This yields the optionality of genitives while
preserving the underlying semantics, as shown in (\ex{1}).  The
association of {\sc gen1} and {\sc gen2} then is determined according
to a hierarchical order of arguments (Bresnan, 1995).

This approach also provides a means of handling certain cases of
categorial shift. For instance, in German temporal and conditional
adjuncts may be realized as PPs dominating an NP headed by a deverbal
noun. English does not have this option, but employs an adjunct-clause
instead. Here, the {\sc gen1} and {\sc gen2} functions of the German
f-structure have to be related correctly to the {\sc subj} and {\sc
  obj} functions of the English f-structure.

\es{\bt{lll}
bei  Karls & Darstellung & des Vorfalls \\
at  Karl-Gen & report & the accident-Gen \\
 mussten & alle & lachen \\
 must-Past & all & laugh \\
\mc{3}{l}{`when Karl reported the accident} \\
\mc{3}{l}{everyone had to laugh'}
\et}

Here the linking of the {\sc gen1} and {\sc gen2} functions to the
appropriate thematic role in the German f-structure drives the
transfer of these functions to the {\sc subj} and {\sc obj} functions
of the English f-structure.

\es{
\[ \left[  \begin{array}{ll}

\hbox{{\sc pred}} & \hbox{`Darstellung'} \\ \\ 

\hbox{{\sc arg-str}} & \left[  \begin{array}{ll} 
\hbox{{\sc arg1}} & \hbox{{\sc agent}} \\
\hbox{{\sc arg2}} & \hbox{{\sc theme}} \\

                 \end{array} \right] \\  

\node{a}{\hbox{{\sc gen1}}} & \left[ \begin{array}{ll} 
\hbox{{\sc pred}} & \hbox{`Karl'} \\
                 \end{array} \right] \\

\node{b}{\hbox{{\sc gen2}}} & \left[ \begin{array}{ll} 
\hbox{{\sc pred}} & \hbox{`Vorfall'} \\
                 \end{array} \right] \\  

\end{array} \right] \]
}

\es{
\[ \left[  \begin{array}{ll}

\hbox{{\sc pred}} & `\hbox{report}<\hbox{{\sc subj}}, \hbox{{\sc obj}}>\mbox{'} \\

\node{c}{\hbox{{\sc subj}}} & \left[ \begin{array}{ll} 
\hbox{{\sc pred}} & \hbox{`Karl'} \\
                 \end{array} \right] \\  

\node{d}{\hbox{{\sc obj}}} & \left[ \begin{array}{ll} 
\hbox{{\sc pred}} & \hbox{`accident'} \\
                 \end{array} \right] \\  

\end{array} \right] \]
}

\nodecurve[l]{a}[l]{c}{1in}[1in]
\nodecurve[l]{b}[l]{d}{1in}[1in]

Under this approach, languages now only differ with respct to the
categorial realisation of the function by case-marked NP or PP.  Thus,
an application of this treatment not only provides an adequate
grammatical analysis of the NP in German, but also facilitates MT.

\section{Conclusion}

This paper has presented innovative approaches for two particular
syntactic phenomena: auxiliaries and multiple genitive NPs.  The
analyses proposed allow the factorization of language particular,
idiosyncratic information.  This results in a cleaner treatment of
auxiliaries by factoring out morphological wellformedness conditions,
and allows for the preservation of argument structure information in
cases like that of the German multiple genitive NP construction, where
syntactically dissimilar constructions express essentially the same
predicate-argument relations. As such, the work presented here can be
seen as a small but necessary step towards the realization of a broad
coverage grammar.  In particular, the feasibility of developing
parallel grammars for differing languages is greatly increased through
the formulation of uniformly applicable, transparent analyses.

\section{Acknowledgments}
We would like to acknowledge Judith Berman, Mark Johnson, Ron Kaplan,
Mar\'{\i}a-Eugenia Ni\~{n}o and Annie Zaenen for the many valuable
discussions that served as input to this paper.

\end{document}